\documentclass[fleqn,10pt]{wlpeerj}
\usepackage{amsmath}

\usepackage[]{algorithm2e}
\usepackage{siunitx}
\usepackage{bm}
\usepackage{url}

\title{A modular software framework for the design and implementation of ptychography algorithms}

\author[1, 2]{Francesco Guzzi}
\author[1]{George Kourousias}
\author[1]{Fulvio Billè}
\author[1]{Roberto Pugliese}
\author[1]{Alessandra Gianoncelli}
\author[2]{Sergio Carrato}
\affil[1]{Elettra Sincrotrone Trieste, Basovizza, Italy}
\affil[2]{University of Trieste, Trieste, Italy}

\corrauthor[1,2]{Francesco Guzzi}{francesco.guzzi@elettra.eu}

\begin{abstract}
Computational methods are driving high impact microscopy techniques such as ptychography. However, the design and implementation of new algorithms is often a laborious process, as many parts of the code are written in close-to-the-hardware programming constructs to speed up the reconstruction. In this paper, we present SciComPty, a new ptychography software framework aiming at simulating ptychography datasets and testing state-of-the-art and new reconstruction algorithms. Despite its simplicity, the software leverages GPU accelerated processing through the PyTorch CUDA interface. This is essential to design new methods that can readily be employed. As an example, we present an improved position refinement method based on Adam and a new version of the rPIE algorithm, adapted for partial coherence setups. Results are shown on both synthetic and real datasets. The software is released as open-source.
\end{abstract}
\begin{document}
\flushbottom
\maketitle
\thispagestyle{empty}
\section*{Introduction}
Today, modern microscopy fully relies on cutting edge computing. The contemporary presence of a large Field Of View (FOV), high resolution and quantitative phase information is the distinctive characteristic of the ptychography diffraction imaging technique \citep{pie2004, Pfeiffer2018}. This phase retrieval scheme is commonly solved through a computationally intensive procedure - \emph{a reconstruction} - which requires algorithms written in a low-level language or following complex HPC paradigms. Writing such programs is typically done by an expert software engineer, and this step may discourage the prototyping or the study of new methods. Ptychography algorithms are complex not only in their implementation, but also \emph{per se}: simple computational systems suffers from setup inconsistencies, which usually takes the form of bad parameter modelling of \emph{positions} \citep{zhang2013}, \emph{distances} \citep{guzziautodiff}, \emph{illumination conditions}  \citep{thibault_2013}, just to enumerate a few. Such problems produce very noticeable artefacts. To improve the quality of the reconstructions, parameter tweaking becomes essential \citep{Reinhardt_2018}, especially for common setups; powerful but even more advanced and complex algorithms \citep{ptypy, Pynx, Marchesini:jo5020} are thus employed.

\subsection*{State of the art}
Iterative algorithms such as ePIE \citep{Maiden2009} or DM \citep{thibault2008} are typically employed for the reconstructions. Recently, the rPIE algorithm \citep{rpie} has been proposed and studied: while being reasonably simple, it provides a fast convergence to a good object estimate \citep{rpie}. Compared to ePIE, we noticed that the rPIE algorithm, at least in our implementation, also provides a large computational FOV, which is comparable to the one seen in more advanced, but also computational eager, optimisation algorithms \citep{NLpaper, Thibault_2012, guzziautodiff}. Indeed, these latter methods are used to refine a previous reconstruction, as they are more prone to stagnation \citep{Thibault_2012}. Being new, the rPIE algorithm is currently lacking of: i) a public implementation; ii) a model decomposition approach; iii) a tested position refinement routine. As a whole, a \emph{recipe} is missing for this kind of algorithm.

\subsection*{Proposed framework - SciComPty}
In this paper we describe how these three elements can be combined within the SciComPty GPU framework, a new software released as open-source \citep{Guzzi2021dataset}, entirely written in the PyTorch \citep{pytorchpaper} Python dialect. Our reconstruction recipe is tested against other solutions and then the results are reported. We designed a fast position refinement technique exploiting Adam \citep{Adam} as a feedback controller. To do so, a fast subpixel registration algorithm \citep{crosscorr2008} has also been implemented via a PyTorch GPU code. This latter element has many uses, for example in CT alignment \citep{guzziCT}, or super-resolution imaging \citep{guarnieri2020, guzzi2018}.The details are described in the "method section". The software and algorithms capabilities are illustrated ("Results section") with reconstructions from simulated and real soft-X-ray data acquired at TwinMic (Elettra) \citep{twinmicstat, twinmic2021}. Within the "Discussion section", we will elaborate more on the features of the reconstructed images. The datasets and the code can be accessed at \citep{Guzzi2021dataset, vuodataset}.

\subsection*{Background}
\begin{figure}[h!]
	\begin{center}
		\includegraphics[width=\linewidth]{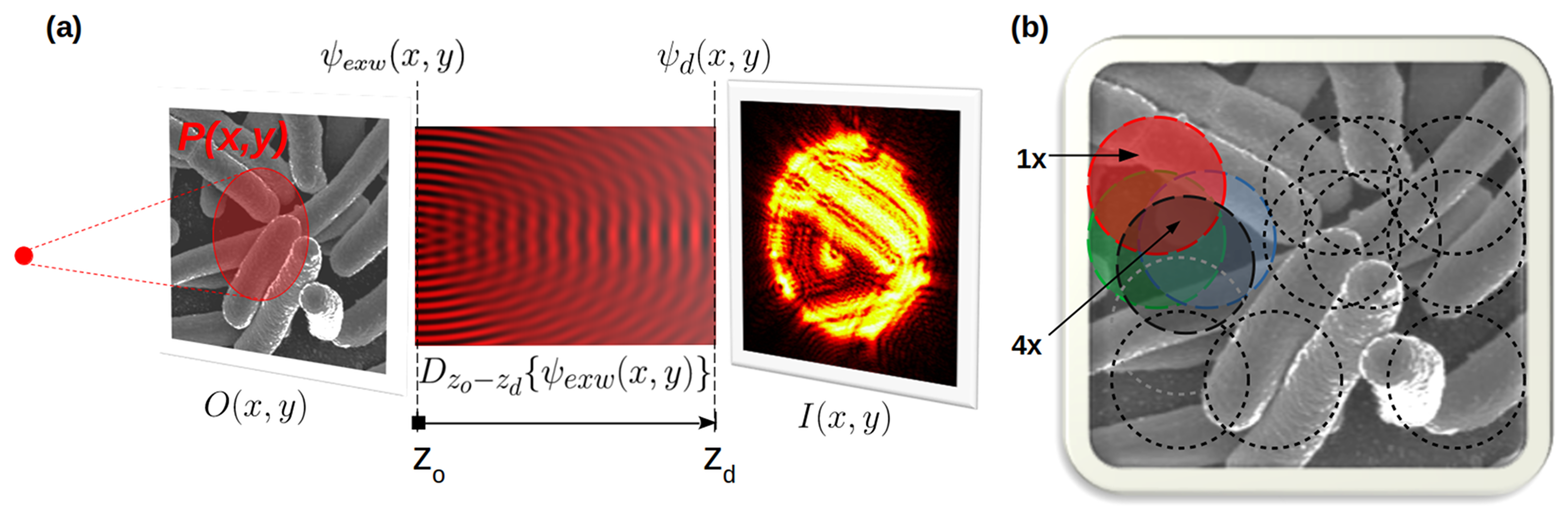}
	\end{center}
	\caption{Panel (a): $P(x,y)$ is shined onto an object $O(x,y)$, producing $\psi_{exw}(x,y)$. At the detector we observe $|\psi_{d}(x,y)|^2$. Panel (b): ptychography uses overlapped illumination (e.g. 4x oversampling) to solve the phasing. The model in Eq. \ref{eq:exw} is repeated for each illuminated region.}
	\label{fig:cdischeme}
\end{figure}

Ptychography aims at recovering the 2D specimen transmission function $O(x,y) \in C$  starting solely from a set of diffraction patterns \citep{Williams2006, Pfeiffer2018}. In a transmission setup (Fig. \ref{fig:cdischeme}), a coherent and monochromatic wavefield $P(x,y) \in C$ is shined onto the specimen ($z=z_O$), placed between the source and the detector ($z=z_d$). In the thin sample approximation \citep{paganin2006}, the \emph{exit-wave} $\psi_{exw}(x,y)$ transmitted by the object becomes:
\begin{equation} \label{eq:exw}
\psi_{exw}(x,y)=\psi(x,y,z_o) = P(x,y)\cdot|O(x,y)|\cdot e^{j\cdot\phi_{O(x,y)}}
\end{equation}

\noindent During the path from the sample to the detector, free-space propagation occurs \citep{paganin2006}, modelled by the $D_z$ operator \citep{jdsim2010}:
\begin{equation} \label{eq:exitdet}
\psi_d(x,y) = \psi(x,y,z_d) = D_{z_d-z_o}\{\psi_{exw}(x,y) \} = D_{z_d-z_o}\{P(x,y,z_o) \cdot O(x,y) \}
\end{equation}

\noindent Finally, a 2D photon detector is used to reveal the intensity (magnitude squared) of the incident radiation, producing a real, discrete, and quantised image $I(x,y)$, which is the diffraction pattern.
\begin{equation} \label{eq:intens}
I(x,y) = \psi_d(x,y) \cdot \psi_d^*(x,y)
\end{equation}

\noindent Ptychography employs translation diversity to over-condition the problem; a large illumination (\emph{probe}) is indeed scanned across the sample, by imposing a large overlap on each illuminated region (Fig. \ref{fig:cdischeme}, panel b); as a result, the object is highly oversampled. A ptychography dataset (real or simulated) is thus composed of multiple diffraction patterns, paired with their corresponding translation coordinates. Geometry and illumination metadata complete the description of a particular instance of the computational forward model.

\noindent During the reconstruction process, the model described by Eq. \ref{eq:intens} must be inverted to gather back the object transmission function $O(x,y)$. Since this inversion cannot, in general, be done analytically, an iterative procedure is employed. Modern versions of the ptychography reconstruction algorithms \citep{thibault2008, Maiden2009} automatically perform the factorisation of the $P(x,y)$ and $O(x,y)$ functions. In a typical imaging-oriented ptychography experiment like ours, the recovery of $P(x,y)$ is just a by-product, though mandatory for the $O$-$P$ factorisation.

\section*{Methods}
\subsection*{Virtual experiment}
Reconstructions apart, SciComPty offers a simple method to perform virtual experiments. To simulate a ptychography dataset, one can implement the general transmission model of Eq. \ref{eq:intens}, and assign particular values to its setup parameters.
In the case of a far-field setup \citep{jdsim2010}, by fixing the detector pixel size $\delta_D$, the corresponding pixel size at the sample plane $\delta_s$ can be readily calculated (Eq. 1 in the supplementary material). The recurring parameters in both simulations and reconstructions are $W$ (detector size in pixels), $S_d$ (detector lateral dimension) and $z_{do}$ (sample-detector distance). In a simulation, $\delta_s$ normally defines the basis for the scanning movements, that in the simplest case follows a grid pattern. Random jitter is added to the $(x, y)$ coordinates, to prevent the \emph{raster scanning pathology} \citep{thibault2008, Dierolf_2010, edo2013}. The overlap factor is defined instead by the maximum step movement. The object function $O(x,y)$ and the illumination function $P(x, y)$ are assembled in magnitude and phase providing two images each. Defining such virtual experiment parameters follows what is actually done during a real experiment (see Listing 1 in the Simulation section of supplementary material).

\subsection*{M-rPIE algorithm} \label{sec:mrpie}
In order to describe partial coherece beams \citep{chenJones2012} the wavefield is decomposed into a set of mutually incoherent probe modes \citep{thibault_2013, odstricil2016, Li16}. For each sample position $(x_j, y_j)$, the $jth$ diffraction pattern $I_j(x,y)$ can be modeled by:
\begin{equation} \label{eq:intensdecomp}
I_j(x,y) = \sum_{p=1}^{M} | D_z \{P_p(x,y) \cdot O(x,y,x_j,y_j) \} | ^2 = \sum_{p=1}^{M} | D_z \{P_p(x,y) \cdot o_{j}(x,y) |^2
\end{equation}

\noindent where $P_p$ with $p \in \{1,...,M\}$ is a particular probe mode; $\bm{o_j} = o_{j}(x,y) = O(x,y,x_j,y_j)$ is the cropped region of the object, corresponding to the \emph{jth} illuminated area in the scan sequence. To reconstruct an object with the model in Eq. \ref{eq:intensdecomp}, we solve for $O(x,y)$ and \emph{a set} of $M$ mutually incoherent probe functions $\bm{P_p} = P_p(x,y)$. A public implementation of rPIE is currently missing and  its multiprobe variant has not been reported yet. We implemented it in SciComPty, calling it M-rPIE. The procedure to design the method follows what has been done for one of the multimode-ePIE (M-ePIE) versions \citep{thibault_2013}, producing the following new update steps (2D arrays depending on $x,y$ are in \textbf{bold}):
\begin{equation} \label{eq:probeupd}
\bm{P_p'}=\bm{P_p}+\alpha_P\frac{(\bm{\psi_{exw_{p,j}}'}-\bm{P_p} \cdot \bm{o_j})\cdot \bm{o_j^{*}} }{\beta\left|\bm{o_j}\right|_{\max}^2+(1-\beta)\left|\bm{o_j}\right|^2};
\end{equation}
\begin{equation} \label{eq:objupdate}
\bm{o'_j}=\bm{o_j}+\alpha_O\frac{\sum_{p=1}^M(\bm{\psi_{exw_{p,j}}'}-\bm{P_p} \cdot \bm{o_j})\cdot \bm{P_p^{*}}}{\gamma(\sum_{p=1}^M|\bm{P_p}|^2)_{\max}+(1-\gamma)\sum_{p=1}^M|\bm{P_p}|^2}
\end{equation}
\noindent As usual, $\bm{o_j'} = o_j'(x,y)$ and  $\bm{P_p'}=P_p'(x,y)$ represent the updated quantities respectively for the current object box and the current probe, while:
\begin{equation}
\bm{\psi_{exw_{p,j}}'}=D_{-z}\{ \sqrt{\bm{I_j}}\frac{\bm{\psi_{exw_{p,j}}}}{\sqrt{\sum_{n=1}^N\left|\bm{\psi_p}\right|^2}} \}
\end{equation}

\noindent is the magnitude-corrected scattering wave produced by the \emph{pth} illumination mode; $\alpha_p$ and $\alpha_o$ represent the update rates of the probe and object estimates and are typically set to $[0.5, 1]$. The role of each denominator is to weight the update where the exit wave is not bright \citep{pie2004, rpie}, reducing thus the noise amplification. $\beta$ and $\gamma$ are regularisation parameters: when set to 1, M-rPIE simplifies to M-ePIE \citep{thibault_2013, rpie}. The net effect of the modifications, weighted by $\beta$ and $\gamma$, is to slow down the update, by introducing a loss cost given by $o_j(x,y)$ and $P_p(x,y)$.
\noindent Increasing the number of modes increases the iteration time by the same factor: that is why it is extremely important the fact that SciComPty is actually working on GPUs.

\subsection*{Adam-based position refinement}
As the resolution increases, the effects of backlash and limited mechanical precision become more and more noticeable. The positions vector is indeed populated by the open-loop commands given to the stage. A correction method must then be introduced \emph{a posteriori}, increasing the reconstruction time. In \citep{NLpaper, guzziautodiff} positions are corrected through an optimisation method, within the gradient-based reconstruction; in \citep{Pynx}, instead, the authors propose to use a gradient-less method based on \citep{Powell} to guide the position refinement procedure.  

\begin{figure}[ht!]
	\begin{center}
		\includegraphics[width=\linewidth]{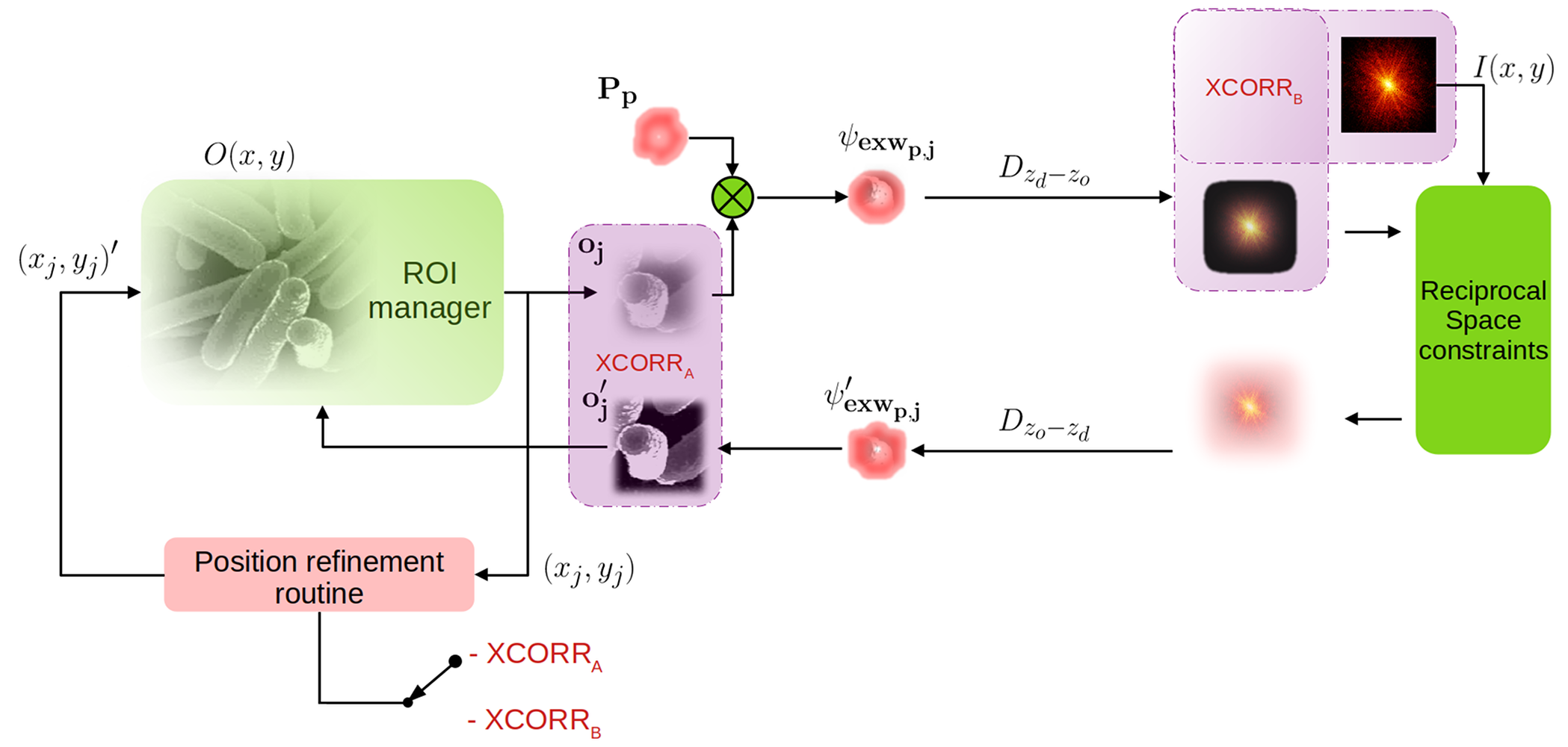}
	\end{center}
	\caption{Position refinement routine integrated into a PIE algorithm; the position error signal (argmax of 2D cross-correlation) can be calculated meaningfully in two points, $XCORR_A$ or $XCORR_B$ (purple boxes). The estimated positions $(x,j)$ are then updated to $(x_j,y_j)'$ iteration by iteration.}
	\label{fig:posrefscheme}
\end{figure}

\noindent In this work, we consider a fast and reliable way to correct translations. Similarly to \citep{zhang2013, Tripathy2014, Loetgering_2015, Dwivedi2018}, it employs a 2D cross correlation signal, calculated at some points in the reconstruction chain. The same principle is used also for CT alignment in \citep{dogatomoalign, guzziCT}, as the synthesised projection $o_j(x,y)$ are inevitably \emph{centred}; when confronted with refined estimates ($o'_j(x,y)$, $XCORR_A$), or measured data ($I(x,y)$, $XCORR_B$), geometrical shifts can then be measured. Figure \ref{fig:posrefscheme} shows how such position refinement scheme can be introduced in the canvas of a PIE reconstruction algorithm: \emph{2D weighted phase correlation} \citep{crosscorr2008} here is used to determine the shift between two different estimates $\bm{o_j}$ and $\bm{o'_j}$ (switch in position $XCORR_A$ in Fig. \ref{fig:posrefscheme}, more details later) belonging to the same crop-box $(x_j, y_j)$.

\noindent In this class of position refinement methods, the error signal (the 2D argmax of the 2D cross-correlation) is extremely small and provides a correction factor that is cumulated at each iteration. A gain factor $\eta_j$ is indeed used to provide a correction. Here we propose to use a \emph{spatially variant} and \emph{adaptive} gain factor, that can change iteration after iteration and is probes-dependent. Our subpixel shift detection is based on the work \citep{crosscorr2008} and its implementation in SciPy \citep{scipy}. We adapted the same algorithm in PyTorch, allowing for fast GPU computation and avoiding costly move operation between the two memory systems, as the core of the reconstruction (M-rPIE) is already working on the GPU. To speed up the procedure for subpixel scales, the matrix multiplication version of the 2D DFT is employed. An upsampled version of the cross-correlation can then be computed within just a neighbourhood of the coarse peak estimate, without the need for zero-padding. Each probe position is updated at each iteration by the following expression:
\begin{equation} \label{eq:posx}
x'_{j}=x_j + \eta_{x,j} \cdot \operatorname*{argmax}_{x} \{XCORR_{i \in [A,B]}\},
\end{equation}
\begin{equation} \label{eq:posy}
y'_{j}=y_j + \eta_{y,j} \cdot \operatorname*{argmax}_{y} \{XCORR_{i \in [A,B]}\},
\end{equation}
\noindent where $\eta_{x,j}$ and $\eta_{y,j}$ are the gain factors $(>>1)$. The variable gain is calculated by using the Adam optimization algorithm \citep{Adam}: Adam is an extension to stochastic gradient descent, and is nowadays widely used in deep learning \citep{8624183}. While in Stochastic Gradient Descent (SGD) the same learning rate is kept constant for all the variables, Adam provides a per-parameter factor that is separately adapted as learning unfolds. This is achieved by considering the evolution of each parameter. In our method, the argmax of the 2D cross-correlation takes the place of the gradient in the Adam algorithm, because its value will maximise the position error gradient. The resulting algorithm is then reported in Alg. \ref{alg:adamalg}:

\begin{algorithm}[H]\label{alg:adamalg}
	\KwData{The cropped object $\mathbf{o_j}$ and its refined estimate $\mathbf{o_j'}$}
	\KwResult{Gain parameters $\eta_{x,j}$ and $\eta_{y,j}$ for the \emph{jth} crop-box (the \emph{jth} position vector $(x,y)_j$)}
	Initialization as in Alg. 1 of \citep{Adam}\;
	\While{reconstruction iteration $>$ 0}{
		t = t+1\;
		$g_t = \operatorname*{argmax}_\mathbf{} \{XCORR_{i \in [A,B]}\}$ \;
		proceed as from row 3 of Alg. 1 of \citep{Adam}\;
		[...]
	}
	\caption{The modified Adam algorithm uses the argmax of the 2D phase correlation as the gradient of the error, then the exponentially damped moving averages are updated as in \citep{Adam}.}
\end{algorithm}

\section*{Results}
\noindent Experiments on simulated and real data were performed. We compared the proposed reconstruction recipe in SciComPty with other state of the art algorithms, also implemented in the advanced \emph{PyNx} software \citep{Pynx, Favre-Nicolin:zy5006}. Simulated data experiments are reported in the supplementary material and confirm that $XCORR_A$ (see Fig. \ref{fig:posrefscheme}) is the best point to estimate the position error. From those results we defined our strategy for the analysis of real datasets.

\subsection*{Synchrotron soft-X-ray experiments}
\noindent Real data have been acquired at the TwinMic beamline of the Elettra Synchrotron facility. We used a 1020 eV (Fig. \ref{fig:cell2} and S1 in supplementary material) and 1495 eV (Fig. \ref{fig:poscorrnocorrreal}) soft-X-ray beam obtained from a secondary source of 15 $\mu$m. The zone plate has a diameter of 600 $\mu m$ with a smallest ring width of 50 nm. At those energies the focus length is respectively of 36 and 24 mm. The sample was placed at $370 \mu $m from the virtual point source, providing a probe size of roughly 9 $\mu m$. The sample-detector distance is set at roughly 75 cm. The Princeton camera is based on a peltier-cooled CCD sensor with a resolution of 1300x1340 pixels, with a pixel size of 20 $\mu$m. 

\begin{figure}[htbp]
	\begin{center}
		\includegraphics[width=\linewidth]{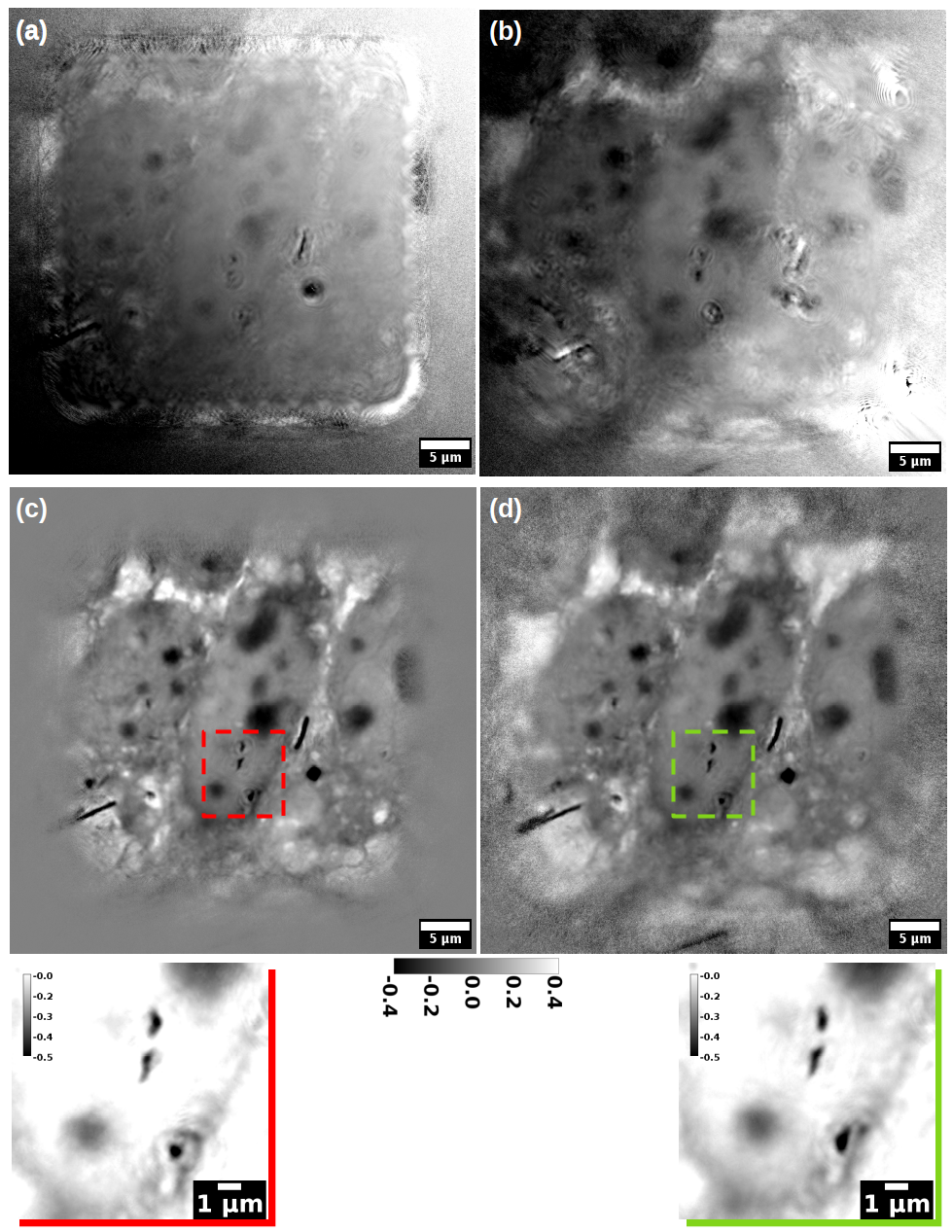}
	\end{center}
	\caption{Reconstruction of a MET cells sample. Panel (a) and (b) show respectively the output of DM and ML, paired with the position correction algorithm \cite{Pynx}. Panel (c) and (d) show the reconstruction with the proposed recipe, using M-ePIE (c) and M-rPIE (d). The insets show how the latter gives fewer ringing artefacts.}
	\label{fig:cell2}
\end{figure}

\noindent Figure \ref{fig:cell2} shows a series of phase reconstructions of a group of chemically fixed Mesenchymal-Epithelial Transition (MET) cells, grown in silicon nitride windows and exposed to asbestos fibres \citep{asbestos2}. The reconstruction in panel (a) has been produced by the DM algorithm, while the one in panel (b) is the output of the ML algorithm \citep{Thibault_2012}. In both, the position refinement method \citep{Pynx} is enabled. In the same configuration, similar results can be obtained by the AP method \citep{MARCHESINI2016815}, that can be seen as a parallelised version of the ePIE algorithm, implemented in PyNx. The reconstruction programs for these images are listed in the supplementary material. Even if stunning, images in panel (a) and (b) appear blurry and full of artefacts. As expected, the ML algorithm provides a better reconstruction than DM; a larger FOV can also be observed. In both, AP was essential to create a meaningful $P(x,y)$, that allowed the object to appear as "reconstructed". At the end of these reconstructions, it was also required to remove a phase modulation; many details about this post-processing can be found in the supplementary material.

\noindent Conversely, panel (c) and (d) show the reconstructions obtained with SciComPty, for the same set of parameters and pre-processed data; in both the two cases, the proposed Adam-based position refinement method is used. Even the simple M-ePIE (panel b) is able to provide a high quality reconstruction, if paired with the proposed position refinement method. Reconstruction quality can be even increased if the proposed M-rPIE method is engaged (panel d): the proposed recipe allows to resolve a large FOV, that becomes not only comparable to the one in panel (b), but in some cases (e.g. in the top left corner) is even larger: the bottom left fibre can be now observed in its entire length, as well as other cell organelles. No post-processing is required at the end of these reconstructions. The reconstruction for a different region of the MET sample can be found in Figure S1 of the supplementary material.

\begin{figure}[htbp]
	\begin{center}
		\includegraphics[width=\linewidth]{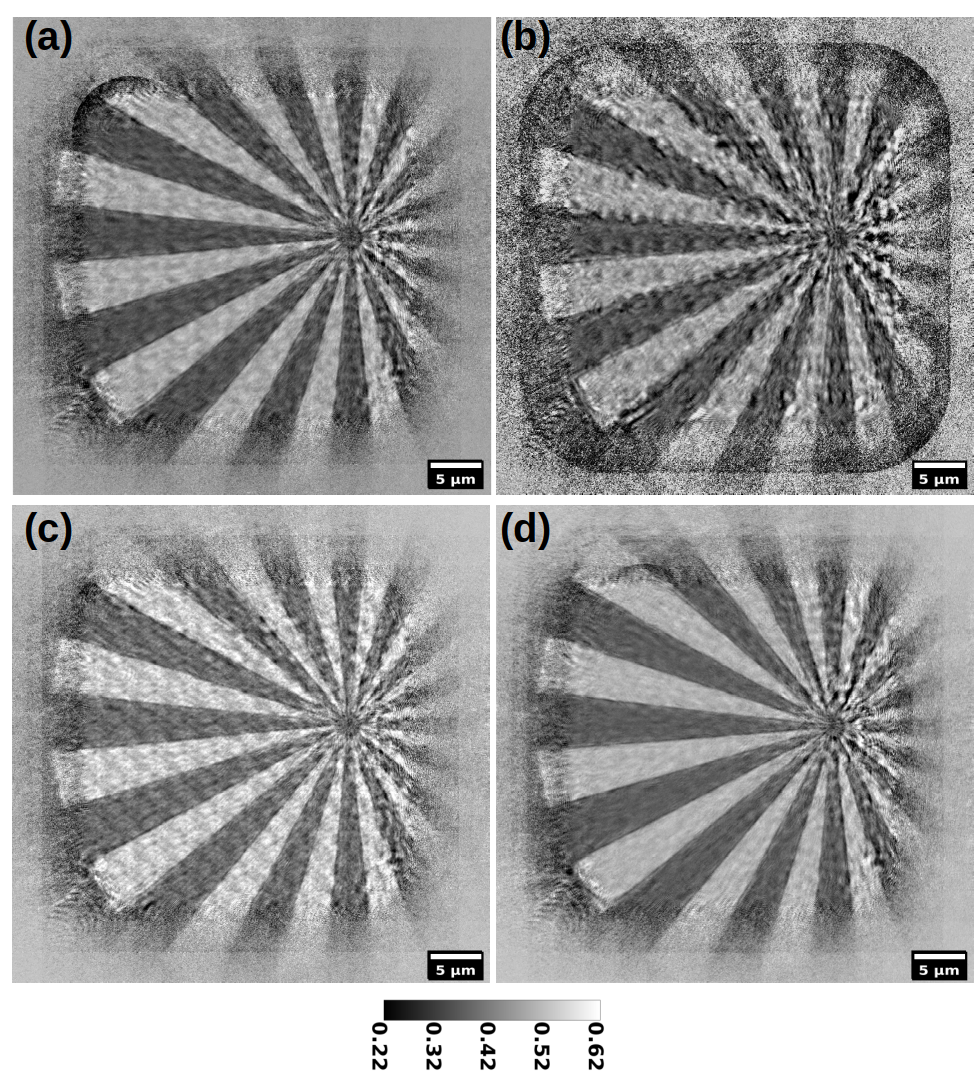}
	\end{center}
	\caption{Reconstruction of a siemens star (magnitude) with different position refinement methods; (a): \citep{zhang2013}; (b): \citep{Pynx}; (c) no correction at all; (d): proposed method.
		\label{fig:poscorrnocorrreal}
	}
\end{figure}

\noindent Figure \ref{fig:poscorrnocorrreal} shows the effects of different position correction methods applied on another real dataset, acquired at 1495 eV. Diffraction patterns are acquired following a regular raster grid. Panel (a) shows the reconstruction with the position refinement algorithm in \citep{zhang2013}; in panel (b) the algorithm in \citep{Pynx} is applied; panel (c) is the reconstruction output with no position correction applied at all, while in panel (d) the proposed Adam-based position correction is applied. As can be seen, even if with simple features, this dataset is quite challenging to correct. While in (a) a form of correction can be seen, the reconstruction in panel (b) is even worse than the one with no correction at all (panel c). The proposed method (d) estimates the best correction, while being fast. Note that in panel (d) the raster grid pathology is quite absent (actual positions are not regularly spaced).

\begin{figure}[ht!]
	\begin{center}
		\includegraphics[width=\linewidth]{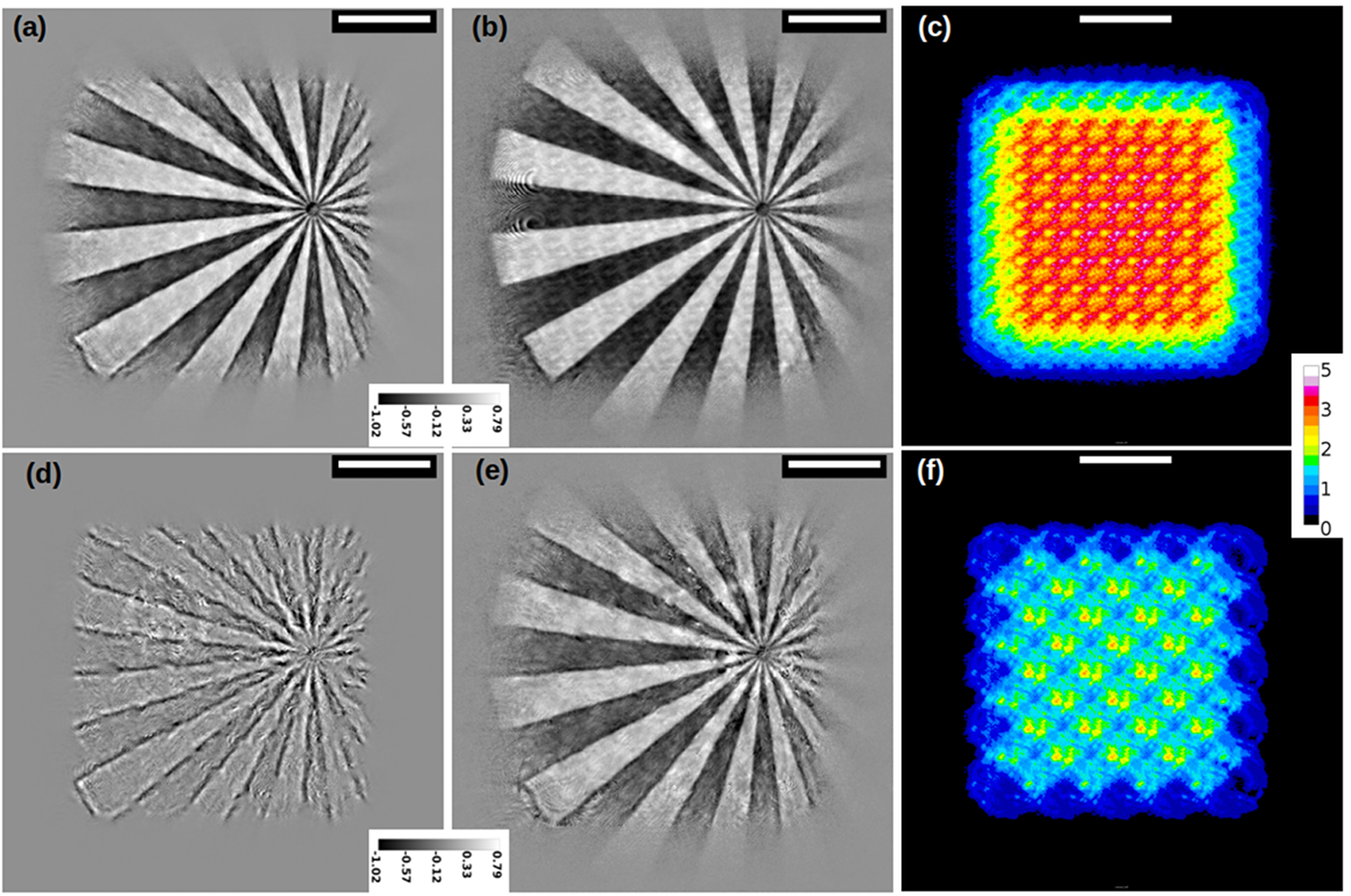}
	\end{center}
	\caption{Reconstruction (phase), obtained using the full dataset (a,b) or a synthetic sparser version (d,e). M-ePIE is used in (a,d), M-rPIE in (b,e). Panel (c) and (f) show the canvas map for each dataset. The white bar represents 10 $\mu$m.
		\label{fig:rpiebetter}
	}
\end{figure}

\noindent A large FOV is the most visible feature in any reconstruction employing M-rPIE (Fig. \ref{fig:cell2}, Fig. \ref{fig:rpiebetter} and Fig. S4). To better analyse this effect, and to disentangle it from the position correction, a sparser dataset has been synthesised from the one of Fig. \ref{fig:poscorrnocorrreal}, producing Fig. \ref{fig:rpiebetter}: panel (a) and (d) are respectively the reconstructions with M-ePIE with the full and a sparse dataset (half of the probes); the canvas map \citep{Kourousias2016a} in panel (c) and (f) shows the two different densities. Panel (b) and (e) show instead the reconstruction with the M-rPIE algorithm: for the same amount of sparsity, the reconstruction is way more resolved in panel (e) than in panel (d).

\subsection*{Implementation details}
\begin{figure}[ht!]
	\begin{center}
		\includegraphics[width=\linewidth]{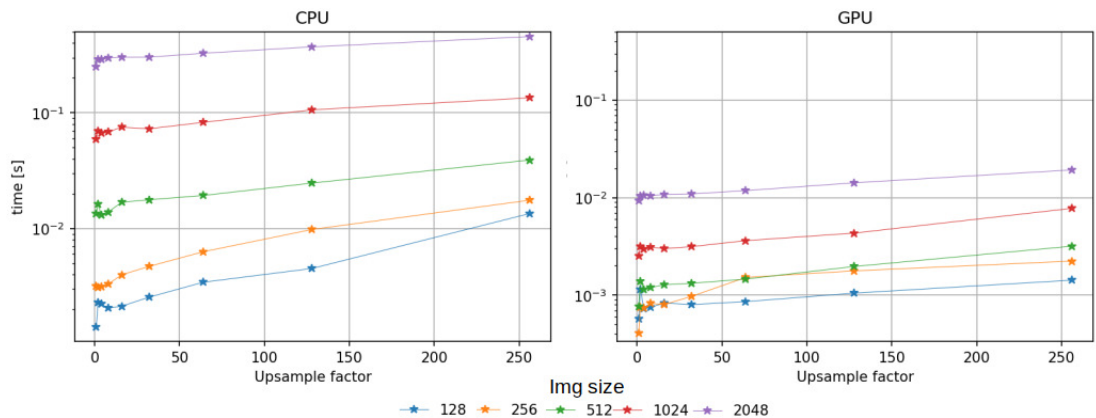}
	\end{center}
	\caption{Benchmark for the position refinement procedure implemented on CPU \citep{scipy} or on GPU (proposed method). The performance gain in terms of speed is about 10x.}
	\label{fig:tempixcorr}
\end{figure}

\noindent SciComPty software framework leverage GPU-based computing for the entire process. The test configuration is reported in the supplementary material. The performance gain obtained by carrying the reconstruction algorithm from the CPU to the GPU is about 10x (0.1 s vs 1.3 s). The reason for such a large gain is the high number of GPU cores which allow to parallelize array calculations. With no position correction, the performance in terms of speed are comparable to other CUDA based solvers implemented in PyNx, which is currently the fastest alternative. Regarding the position correction, at the best of our knowledge, no implementation of \citep{crosscorr2008} is readily available for GPU computing; consequently, we implemented it in PyTorch GPU. Having part of the algorithm on the GPU (reconstruction) and part on the CPU (registration) would have been detrimental, due to the overhead given by the data transfer. It is indeed extremely important that all the required arrays reside in the same domain, minimising copy operations.
Figure \ref{fig:tempixcorr} shows the computation time measured for the registration algorithm implemented on the two devices: the performance gain is of about an order of magnitude. The acceleration is significant, as during the reconstruction the same operation has to be performed for any position in the dataset.

\section*{Discussion}
\noindent A larger FOV is the most noticeable effect in the reconstructions obtained with M-rPIE (see Fig. \ref{fig:cell2}d and Fig. S4). A well-designed regularisation is considered to be the main candidate for a successful phase-retrieval, that is shown to work better for under-sampled areas (Fig. \ref{fig:rpiebetter}). This effects automatically translates in a better reconstruction in the case of sparser scanning, as the higher resolving power can effectively cope with the missing information between different subsequent measures. Sparse scanning means reduced acquisition and reconstruction times, as well as lower radiation dose, and that is why the implementation of the rPIE algorithm in both its single and multi-probe versions is essential for a modern beamline. The fact that these algorithms are implemented on GPU, in a simple way and using open-source tools, effectively allows to use these algorithms.

\noindent The positioning error induced naturally by the sample stage mitigates the raster grid pathology (see Fig. \ref{fig:poscorrnocorrreal}), as the actual movements performed by the motors differs from the position commands sent by the control software. Indeed, as backlash, irregular friction and limited mechanical precision introduce a small irregular jitter, it is not required to add it during a scan. Raster grid artefacts, howewer, \emph{are not} an effect due to position error; the pathology exists and is a consequence of sampling, as they appears also for perfectly ideal simulations with regular grids (see supplementary material). A position refinement method applied \emph{a posteriori} may greatly reduce the artefact, if the actual positions are not regular. That is why a fast implementation of a position refinement algorithm (as deployed by this work) is essential. The Powell method in \citep{Pynx} tends to fail, as the loss function may be incorrectly estimated by the optimiser. Two are the supposed main characteristics which describe the performances of the Adam-based method: i) being based on the evolution of the error through many iterations, the gain estimation is more robust than a simple look-back at the previous iteration; ii) different regions of the sample may require different gains, that have to be adjusted separately.

\noindent In CT alignment, the position error estimation is carried out at position $XCORR_B$ (see Fig. \ref{fig:posrefscheme}). In supplementary material we briefly analysed the dynamics of this kind of error signal, concluding that this configuration performs poorly in a ptychograpy environment, compared to the proposed $XCORR_A$: the use of complex quantities provides a more robust estimate than its real-only counterpart; in addiction, images at the object plane tend to be way more intercorrelated than the one at the detector plane, reducing the noise in the cross-correlation.

\section*{Conclusion} \label{sec:concl}
In this paper, we presented solutions for three major flaws in ptychography: complex software architectures, partial coherence and position errors. By combining together the proposed solutions, we provide a recipe that is giving good results in many ptychography experiments performed at synchrotron and FEL laboratories (Elettra Sincrotrone Trieste). The development of a multi-mode variant for rPIE is essential to reduce the overlap condition, allowing for sparser measurements and thus providing shorter acquisition and reconstruction time. This is critical for dynamics experiments (e.g. with many energies and pump-probe delays). Reducing the lag between measurements and reconstruction is a critical research path, which has a significant impact on all the fields that use this kind of microscopy. We designed these solutions by employing our new GPU-accelerated ptychography software framework, SciComPty, which can be used to easily study and develop both state-of-the-art and new reconstruction algorithms. Throughout the text, we presented several reconstruction examples from real and synthetic datasets, comparing them to other state-of-the art solutions. The position refinement procedure relies on a fast subpixel registration algorithm that also runs on GPU. The entire software is provided to the research community as open source and can be downloaded from \citep{vuodataset, Guzzi2021dataset}.

\section*{Acknowledgements}
We are grateful to Roberto Borghes for his fundamental work on the TwinMic microscope control system and to Iztok Gregori for his work on the HPC solution. 

\section*{Author contributions statement}

GK, FB, RP, AG, SC started the project, GK, AG, FG, SC, FB conceived the experiments, FG, GK, SC, FB designed the method, AG, GK supervised the experiments. All authors participated in the experiments and contributed to the manuscript.

\section*{Additional information}

The authors declare no conflict of interest.

\bibliography{bibliography.bib}

\end{document}